\documentclass[12pt]{article}
\usepackage{amssymb,graphicx}
\usepackage{amsmath}
\usepackage{natbib}

\newcommand{\cip}{\mbox{\,$\perp\!\!\!\perp$\,}}
\newcommand{\ncip}{\mbox{\,$\perp\!\!\!\perp$\hspace{-0.3cm}/\hspace{0.2cm}\,}}
\newcommand{\cd}{\mbox{$\,|\,$}}

\newcommand{\rarr}{\mbox{$\cdots\!\triangleright$}}
\newcommand{\larr}{\mbox{$\triangleleft\!\cdots$}}
\newcommand{\p}{\mathbb{P}}
\newcommand{\E}{\mathbb{E}}
\newcommand{\squar}[5][20]{
	\put(#2,#4){\line(1,0){#1}}
	\put(#2,#4){\line(0,1){#1}}
	\put(#3,#5){\line(-1,0){#1}}
	\put(#3,#5){\line(0,-1){#1}}}

\title{Connecting actuarial judgment to probabilistic learning techniques with graph theory}
\author{Roland Ramsahai}

\begin{document}
\maketitle

\begin{abstract}
Graphical models have been widely used in applications ranging from medical expert systems to natural language processing. Their popularity partly arises since they are intuitive representations of complex inter-dependencies among variables with efficient algorithms for performing computationally intensive inference in high-dimensional models. It is argued that the formalism is very useful for applications in the modelling of non-life insurance claims data. It is also shown that actuarial models in current practice can be expressed graphically to exploit the advantages of the approach. More general models are proposed within the framework to demonstrate the potential use of graphical models for probabilistic learning with telematics and other dynamic actuarial data. The discussion also demonstrates throughout that the intuitive nature of the models allows the inclusion of qualitative knowledge or actuarial judgment in analyses.
\end{abstract}

\section{Introduction}
The aim of improvements in data driven exercises in insurance has led to the desire to gather additional data  than traditionally available. In addition to underwriting characteristics such as age, gender and address, technology now allows the collection of many more variables. Examples include dynamic data from sensors for driving behaviour in vehicles, appliance and electrical usage in homes and static data from external databases on traffic violations, crime scores or credit scores. High dimensional models arise if modelling sensor data at multiple time points and the individual variables that comprise summary scores. Reasoning with a large number of variables can become unnecessarily complex without any actuarial judgment. For example, it may not be necessary to include hundreds of rating factors as predictors if many of them are known to be related or unnecessary.

This discussion proposes the use of graph theory as a means of translating intuitive reasoning to mathematical properties. This is done via graphical models, which involve the use of graph theory to formulate probabilistic models \citep{L:1996}. The approach has been used in applications such as medical expert systems \citep{FSMB:1989}, natural language processing \citep{BNJ:2003}, image processing, bioinformatics and others \citep{WJ:2008}. Graphical models provide an intuitive representation of the structure or network of inter-relationships among variables, both hidden and observable. This is exploited in `deep' learning techniques, where complex hidden structures are postulated to explain features of the observable part of the network. Another advantage of graphical models is that they allow inference and learning with efficient algorithms for complex computations \citep{J:2004}. This is quite important with models involving large datasets and many variables.

A graphical model consists of a qualitative and a quantitative component. The qualitative structure is a pictorial representation of which variables are related. An arrow between two variables represents a direct influence or dependence and is called a `directed edge'. They can be specified before beginning any discussion on the quantitative aspects of the model and are helpful for making actuarial judgments or communicating relevance and irrelevance without the complexities of any probabilistic modelling. After the qualitative picture is clear, missing links or connections are interpreted in probabilistic terminology as independence between variables. The implied independence between variables is a property of their joint distribution and provides a bridge from the network structure to the quantitative aspect of the model.

\begin{figure}\centering
 \begin{picture}(180,20)
   \put(10,10){\circle{20}}\put(6,7){B}
   \put(20,10){\vector(1,0){30}}
   \put(60,10){\circle{20}}\put(56,7){P}
   \put(70,10){\vector(1,0){30}}
   \put(110,10){\circle{20}}\put(106,7){C}
	\put(160,10){\circle{20}}\put(156,7){S}
	\put(150,10){\vector(-1,0){30}}
\end{picture}
\caption{Graphical model relating commercial auto claims $C$, average daily amount of time parked $P$, industry classification $B$ and maximum speed $S$.}	\label{fig:carinsex}
\end{figure}
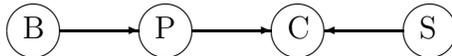

Consider the example in Figure~\ref{fig:carinsex}, which relates the total annual claims for a single vehicle commercial auto policy $C$, average daily amount of time for which the vehicle is parked $P$, industry classification of the business $B$ and maximum speed of the vehicle $S$.
The model arises from the assumption that the actual business classification is only helpful for claim prediction as a proxy for the number of hours that the vehicle is not in use or parked. If data is available from sensors that can detect when the vehicle is parked then business class is no longer relevant. The maximum speed of the driver is however of predictive importance, regardless of whether the other factors are in the model. These assumptions are easily obtained from Figure~\ref{fig:carinsex}, which shows a direct edge from $S$ to $C$ but only edges from $B$ to $C$ that are via $P$. The model assumption is encoded by omitting any direct edge from $B$ to $C$, without any consideration of the actual probability values or even the state space of the variables.

The existence of relationships between variables can be informed by knowledge on causal connections but causality is intentionally omitted from this discussion. It requires additional semantics for the graphs and probability distributions. The focus here is on predictive relevance or irrelevance as defined by dependence or independence. The models of irrelevance allow large complex models to be sub-divided into smaller simpler sub-models. This is helpful for model specification since local relationships between subsets of variables are easier to focus on before aggregation into a model of global relationships. Probability calculations for the model also benefit from this modular property. Data on one variable in a sub-model can be propagated within the sub-model and further throughout the entire global model to make predictions about all variables. The computations required for this updating process is slow with many variables but algorithms exploit the modular property in large graphs to improve its efficiency.

Section~\ref{sec:hiexamp} demonstrates the use of the framework with a simple model of property damage claims for a home insurance policy. Section~\ref{sec:grmod} introduces explains the uses of various algorithms in the literature for predictions and model selection with graphical models. Section~\ref{sec:actgrmod} presents multiple examples from current actuarial practice that could benefit from being expressed as graphical models. Section~\ref{sec:dynpred} introduces the concept of dynamic models and Section~\ref{sec:firemod} extends it to the case of models with both static and dynamic variables. Section~\ref{sec:dynhomeex} describes models for property damage claims in a smart home with dynamic and static variables and structures of unobservable variables. The directed graphs used throughout are often referred to as `Bayesian networks' but their use extends well into frequentist territory.

\section{Home insurance property damage example}\label{sec:hiexamp}
This section describes various graphical models to predict the total amount of property damage claims $C$ on a home insurance policy. Data is available on whether the house construction is non-combustible ($K=1$) or not ($K=0$), and whether the house has a front door burglar alarm ($D=1$) or not ($D=0$). Predictions are done by specifying the model structure and then querying it to obtain predictions. The model structure is represented by a directed graph, which consists of a set of points or nodes that are connected by directed edges or arrows. Each arrow represents a direct connection between a pair of nodes but nodes may be indirectly related to each other by multiple arrows via intermediary nodes. A very simple model is used here to intuitively explain the approach but much larger models with many variables are required in practice, further emphasizing the need for the efficient algorithms that are available with graphical models.

\subsection{Specify model structure with relationships and probabilities}
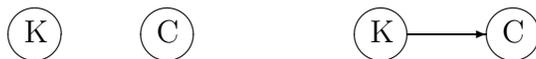
\begin{figure}[htbp]
	\centering
		\begin{picture}(70,30)
		\put(10,10){\circle{20}}\put(06,7){K}
		\put(60,10){\circle{20}}\put(56,7){C}
	\end{picture}\hspace{2cm}
	\begin{picture}(70,30)
		\put(10,10){\circle{20}}\put(06,7){K}
		\put(20,10){\vector(1,0){30}}
		\put(60,10){\circle{20}}\put(56,7){C}
	\end{picture}
\caption{Models with construction $K$ irrelevant (left) and relevant (right) to claims $C$.}
\label{fig:Exlon}
\end{figure}

Consider the case where it is assumed that property damage losses are primarily flood related and thus unrelated to construction class. This is represented by the graphical model in Figure~\ref{fig:Exlon} (left) with no arrow from $K$ to $C$. It is said that $C$ is independent of $K$ and the chance of a loss is the same regardless of the construction class. This lack of relationship can be written as $\p(C\cd K=k)=\p(C)$ for $k=0,1$, without specifying the actual values of the probabilities. The conditional probability $\p(C\cd K)$ expresses the dependency of $C$ on $K$.

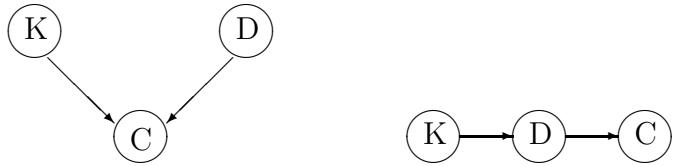
\begin{figure}[htbp]
	\centering
\begin{picture}(90,60)
  \put(10,50){\circle{20}}\put(6,47){K}
  \put(15,40){\vector(1,-1){25}}
  \put(90,50){\circle{20}}\put(86,47){D}
  \put(85,40){\vector(-1,-1){25}}
  \put(50,10){\circle{20}}\put(46,5){C}
\end{picture}\hspace{2cm}
\begin{picture}(100,20)
  \put(10,10){\circle{20}}\put(6,7){K}
  \put(20,10){\vector(1,0){20}}
  \put(50,10){\circle{20}}\put(46,7){D}
  \put(60,10){\vector(1,0){20}}
  \put(90,10){\circle{20}}\put(86,7){C}
\end{picture}
\caption{Graphical models of the relationship between $C$, $K$ and $D$.}
\label{fig:ExlonD}
\end{figure}

Alternatively, property damage losses may be from fire outbreaks and vary with construction. This is represented by Figure~\ref{fig:Exlon} (right), which includes an arrow from $K$ to $C$. It is said that $C$ is not independent of $K$ and written as $\p(C\cd K=k)\neq\p(C)$ for $k=0,1$. The presence of a relationship is specified without actual probability values. If the model is extended to include $D$ then it may be assumed that losses from property damage are related to both construction and crime protection. This implies that $C$ is dependent on both $K$ and $D$ and can be represented by Figure~\ref{fig:ExlonD} (left) with arrows from $K$ and $D$ to $C$.

Alternatively, it may be that claims are primarily crime related and non-combustible construction is more common in modern homes that are more likely to have a door alarm. In this case, $K$ is only related to claims via its predictive door alarm relationship. This is represented by Figure~\ref{fig:ExlonD} (right). It is said that $C$ is independent of $K$ given $D$ and the relationship is expressed as $\p(C\cd D,K)=\p(C\cd D)$. This is written as $C\cip K\cd D$, using the popular notation of \citet{D:1979}. 

After defining the relationships in each case, the model specification is completed by defining the conditional probability values. This is done by specifying the distribution of each variable conditional on its influences. Figure~\ref{fig:Exlon} (left) requires $\p(K)$ and $\p(C)$ since all variables are unrelated. Figure~\ref{fig:Exlon} (right) requires $\p(K)$ and $\p(C\cd K)$ since $K$ is relevant to $C$. Figure~\ref{fig:ExlonD} (left) requires $\p(C\cd K,D)$, $\p(K)$ and $\p(D)$. Figure~\ref{fig:ExlonD} (right) requires $\p(C\cd D)$, $\p(D\cd K)$ and $\p(K)$.

\subsection{Extracting predictions by updating evidence}
The model consists of a structure of relationships and certain conditional distributions. These can be used to compute any marginal or conditional distributions for the purpose of prediction. Predictive models essentially provide the conditional distribution of an outcome given its predictors so that data on the predictors can be plugged into the model to forecast the outcome. For the example in Figure~\ref{fig:Exlon} (right), imposing a generalized linear model relating $C$ and $K$ provides the conditional distribution $C\cd K\sim\mathcal{D}\{g^{-1}(\beta K+c)\}$ for prediction, where $g(\cdot)$ is the link function and $\mathcal{D}(\cdot)$ is the outcome distribution. Without data on the predictor $K$, forecasts are based on the marginal distribution $\p(C)$. This is the chance of a claim without knowledge or evidence on the construction class $K$. The distribution $\p(C)$ can be computed using iterated expectations since $\p(K)$ is available from the model definition.

In the case where data on the construction class of a home is available and it is observed to be non-combustible then the distribution of claim amount is best expressed by the conditional distribution $\p(C\cd K=1)$. This quantity is most appropriate for forecasting since it incorporates the available knowledge or data. It can be obtained by plugging in $K=1$ into the predictive model that defines the relation between $C$ and $K$. The shift from predictions based on $\p(C)$ to $\p(C\cd K=1)$ is referred to as an update based on evidence on $K$. For the example in Figure~\ref{fig:Exlon} (left), the update makes no difference since $K$ is irrelevant to $C$ and $\p(C)=\p(C\cd K=1)$. Updating predictions on $C$ from evidence on $K$ is straightforward in Figure~\ref{fig:Exlon} (right) since it involves evidence on a direct influence. The convenience arises because distributions conditional on direct influences are pre-specified in formulating the model and are available to simply plug in any observed evidence.

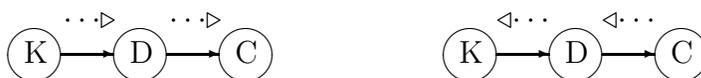
\begin{figure}[htbp]
	\centering
\begin{picture}(100,30)
  \put(10,10){\circle{20}}\put(6,7){K}
  \put(20,10){\vector(1,0){20}}
  \put(50,10){\circle{20}}\put(46,7){D}
  \put(60,10){\vector(1,0){20}}
  \put(90,10){\circle{20}}\put(86,7){C}
	\put(21,20){$\rarr$}\put(61,20){$\rarr$}
\end{picture}
\hspace{2cm}
\begin{picture}(100,30)
  \put(10,10){\circle{20}}\put(6,7){K}
  \put(20,10){\vector(1,0){20}}
  \put(50,10){\circle{20}}\put(46,7){D}
  \put(60,10){\vector(1,0){20}}
  \put(90,10){\circle{20}}\put(86,7){C}
	\put(20,20){$\larr$}\put(60,20){$\larr$}
\end{picture}
\caption{Efficient update of evidence with local simple computations instead of complex global computations.}
\label{fig:EffComp}
\end{figure}

More generally, it is possible to update predictions given evidence on variables which are not direct influences. A simple example is the use of Bayes' theorem for the reverse updating of the prediction of $K$ with evidence on $C$ by computing $\p(K\cd C)$ for the model in Figure~\ref{fig:Exlon} (right). Algorithms exist for extending Bayes' theorem to the general case of updating predictions with evidence from anywhere in the network. They exploit the structure of relationships to improve efficiency. For example, predictions on $C$ can be efficiently updated with evidence on $K$ in Figure~\ref{fig:ExlonD} (right) by updating $D$ from $K$ and then updating $C$ from $D$. The computation steps are represented by the dashed arrows in Figure~\ref{fig:EffComp} (left). The dashed arrows represent steps or a flow of evidence for this particular computation, as opposed to the solid arrows that represent assumptions which actually constitute the model. The efficiency in this case is achieved by exploiting the network of relationships, which entail a lack of direct influence from $K$ to $C$.

The reverse process of updating predictions on $K$ with evidence on $C$ is represented in Figure~\ref{fig:EffComp} (right). Both computations in Figure~\ref{fig:EffComp} involve sub-dividing a global complex computation into simpler local computations by exploiting the modular nature of the graphical models. Global predictions about any variable can be updated from evidence on any other variable, regardless of the direction of influence between them or whether they belong to the same local sub-model. Efficiency in computations is particularly important when there are large models with many variables.

\section{Practical use of graphical probabilistic models}\label{sec:grmod}
The graphical structure uses nodes to represent random variables and directed edges or arrows to represent assumptions about dependence between variables. This graphical representation has been used to develop algorithms that exploit graph theoretical results for inference with probabilistic models. Probabilistic inference involves the computation of conditional distributions given evidence or observations from other variables. This is evidence or belief propagation can be impractical with many variables but the junction tree algorithm \citep{JLO:1990} can perform the exact computation efficiently in certain models. It exploits the graph theoretical properties by transforming the graphical structure and performing local computations between connected nodes, referred to as `message passing' \citep{P:1982}. For certain models, it is necessary to use `loopy' belief propagation or variational algorithms which perform approximate computations. These approximations have been successful in practice and an overview of them is given in \citet{J:2004}. 

The omission of edges between certain pairs of nodes in the graphical structure indicates that there is no direct connection between those nodes. These omitted edges essentially result in nodes being separated and represent conditional independence assumptions about the probability distribution of the variables. The identification of separated variables or conditional independent variables from a graph is important for making computations efficient and intuitively interpreting model assumptions. In general, each variable is independent of its `non-descendants' given its `parents', where parents are its direct influences and non-descendants are those variables which are not influenced directly or indirectly via other variables. More complex conditional independence relations are implied by the properties of the graph and can be derived using the `d-separation' criterion of \citet{VP:1988} or the `moralization' criterion of \citet{LDLL:1990}.

This discussion focuses on the use of graphical models whose qualitative structure is chosen by actuarial judgment and gives little detail on learning the conditional probabilities. However it is possible to learn both the structure of interconnections and the parameters of the conditional probability distributions from data. The former is done as a first step and the latter is done as a second step with efficient algorithms that exploit the graphical structure learnt in the first step. The second step may involve maximum likelihood or Bayesian estimation. Algorithms for estimation are described in \citet{H:1998}.

Structure learning usually involves score based or constraint based methods. The score based methods assign a score to various graphical structures, such as a penalized log-likelihood to avoid over-fitting, and select the one that optimizes the score. Constraint based methods perform tests of conditional independence relations and select the structure which represents the collection of relations detected in the data. \citet{S:2010} describes both types of algorithms and their practical implementation. Variable selection is inherent in structure learning since variables that are separated in the chosen structure would be deemed as irrelevant for inference and ignored.



\section{Graphical models for actuarial techniques}\label{sec:actgrmod}

\subsection{Frequency-severity models}
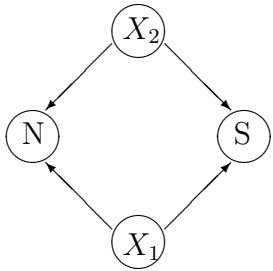
\begin{figure}[htbp]
	\centering
	\begin{picture}(180,100)
	\put(40,50){\circle{20}}\put(36,47){N}
	\put(70,85){\vector(-1,-1){25}}
  \put(80,90){\circle{20}}\put(74,87){$X_2$}
  \put(90,85){\vector(1,-1){25}}
  \put(120,50){\circle{20}}\put(116,47){S}
  \put(70,15){\vector(-1,1){25}}\put(90,15){\vector(1,1){25}}
  \put(80,10){\circle{20}}\put(74,5){$X_1$}
\end{picture}
\caption{Model in which explanatory variables $X_1$ and $X_2$ can be used to model frequency and severity separately.}
\label{fig:frqsev}
\end{figure}

Since insurance losses occur at random points in time, the total loss is usually assumed to follow a compound distribution of identically distributed individual losses. Let the total claims in a period be $C$, the number or frequency of losses be $N$ and the average severity of losses be $S=C/N$. While it is clear that the total claims is the product of frequency and severity, $C=S\times N$, the expected total claims is given by $\E(C)=\E(N)\E(S\cd N)$. As highlighted by \citet{GGS:2016}, traditional models for non-life insurance claims assume that frequency and severity are independent to justify that the expected total claims is the product of the expected frequency and expected severity. This assumption is written as $S\cip N$ and implies that $\E(S\cd N)=\E(S)$ and hence $\E(C)=\E(S)\E(N)$. The same idea can be extended to the case where there are explanatory variables. For simplicity, consider the extension to the case where there are two explanatory factors $X_1$ and $X_2$ for predicting the total claims $C$. By iterated expectations, it can be shown that
\begin{align}
 \E(C\cd X_1,X_2)=\E(N\cd X_1,X_2)\E(S\cd N,X_1,X_2).	\label{eq:frsev}
\end{align}
The first component $\E(N\cd X_1,X_2)$ in the factorization in Eq.~(\ref{eq:frsev}) is the parameter of interest in a frequency model. The second component $\E(S\cd N,X_1,X_2)$ in the factorization in Eq.~(\ref{eq:frsev}) is not the typical parameter of interest in a severity model since it includes frequency as a predictor. It can only be simplified to the usual severity model parameter $\E(S\cd X_1,X_2)$ under the assumption that $S\cip N\cd (X_1,X_2)$, as represented in Figure~\ref{fig:frqsev}.

Figure~\ref{fig:frqsev} could be justified intuitively using qualitative arguments. It may be reasoned that the explanatory factors influence both $N$ and $S$ but capture all of their dependence. The key property is that there is no link between $S$ and $N$, other than via $(X_1,X_2)$. Both rating factors must be included in the severity model for an unbiased analysis since $S$ is not independent of $N$ given $X_1$ or $X_2$ alone, written as $S\ncip N\cd X_1$ and $S\ncip N\cd X_2$. Predictors may be eliminated if further assumptions are justified by expert opinion or data driven model selection. For example, if it is assumed that $X_2\cip N$ then the arrow from $X_2$ to $N$ in Figure~\ref{fig:frqsev} would be removed. This would imply that $X_2$ could be removed from the frequency model. It could also be removed from the severity model since it would no longer be needed to separate $N$ and $S$.

Explicitly discussing the assumption $S\cip N\cd (X_1,X_2)$ fosters transparency because the approach is invalid if the assumption is violated. For example, rating a professional liability product with profession $X_1$ and revenue $X_2$ may be inadequate if claims are more severe for professions where they are more frequent. This would suggest there is a direct arrow from $N$ to $S$ in Figure~\ref{fig:frqsev} and severity must be modelled with frequency. It may also be that the direct relation between $S$ and $N$ is via some additional rating factor linked to regulations or economic factors not captured by revenue. The classic frequency-severity approach would be justified if the missing factor is included as a rating factor. Multiple models which allow for dependence between frequency and severity are given in \citet{GGS:2016} and its references.

Similar issues arise using frequency-severity methods for developing ultimate losses in reserving. If $X_1$ is policy year and $X_2$ is development year of a claim, frequency and severity triangles can only be developed separately if $S\cip N\cd (X_1,X_2)$. For example, this could be violated with professional liability claims if policy years with more severe claims tend to have more frequent claims. For any given policy year, since the development pattern is different for large and small claims then the development of the severity of claims would vary with frequency.


\subsection{Risk factors in generalized linear models}
\begin{figure}[htbp]
	\centering
\begin{picture}(100,60)
	\put(10,50){\circle{20}}\put(4,47){X$_1$}
	\put(50,50){\circle{20}}\put(44,47){X$_2$}
  \put(90,50){\circle{20}}\put(84,47){X$_3$}
  \put(15,40){\vector(1,-1){25}}\put(85,40){\vector(-1,-1){25}}\put(50,40){\vector(0,-1){20}}
  \put(50,10){\circle{20}}\put(46,5){Y}
\end{picture}\hspace{1cm}
\begin{picture}(100,60)
	\put(10,50){\circle{20}}\put(4,47){X$_1$}
	\put(50,50){\circle{20}}\put(44,47){X$_2$}
  \put(90,50){\circle{20}}\put(84,47){X$_3$}
  \put(15,40){\vector(1,-1){25}}\put(80,50){\vector(-1,0){20}}\put(50,40){\vector(0,-1){20}}
  \put(50,10){\circle{20}}\put(46,5){Y}
\end{picture}
\caption{Models relating outcome $Y$ to explanatory variables ($X_1$,$X_2$,$X_3$), where all (left) or some (right) of the explanatory variables are relevant.}
\label{fig:genlim}
\end{figure}
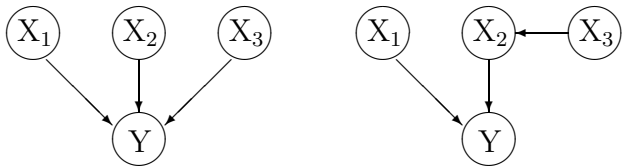

In actuarial practice, generalized linear models are used for pricing and stochastic reserving. The outcome of interest, usually claim frequency or severity, is defined as having some relationship to explanatory variables. These explanatory variables can be risk characteristics for pricing and accident or development year for reserving. The relationship between the outcome and explanatory variables is stochastic since parameters of the outcome distribution are expressed as a function of the explanatory variables. Consider an example of a model in which the outcome $Y$ is related to the explanatory variables $(X_1,X_2,X_3)$ via the function $\mu_{Y|X}=g(X_1,X_2,X_3)$, where $\mu_{Y|X}=\E(Y\cd X_1,X_2,X_3)$. All of the explanatory variables have a predictive influence on $Y$ and are usually assumed to be independent of one another, written as $X_1\cip X_2\cip X_3$. This can be represented by Figure~\ref{fig:genlim} (left).

Alternatively, it may be that $X_3$ is correlated with $X_2$ and unnecessary if $X_2$ is included in the model. This can be written as $X_3\ncip X_2$ and $Y\cip X_3\cd X_2$ and represented by Figure~\ref{fig:genlim} (right). The form of the relationship between outcome and explanatory variables will then be adjusted to $\mu_{Y|X}=g(X_1,X_2)$. Such selection of predictors can be done with qualitative expert judgment before any technical considerations. This is because the network does not express the actual form of the relationship between variables, just which ones are related. The graph remains the same regardless of the link function, the form of the linear predictor or the outcome distribution eventually chosen to define $\p(Y\cd X_1,X_2,X_3)$. Qualitative refinement of predictors still allows additional selection at the quantitative stage using methods such as statistical tests of significance for coefficients.

\subsection{Predictive models with summary scores}
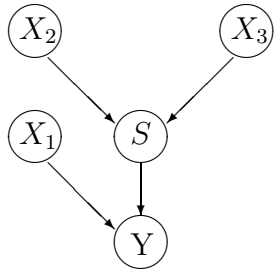
\begin{figure}[htbp]
	\centering
\begin{picture}(100,100)
	\put(10,90){\circle{20}}\put(4,87){$X_2$}
	\put(15,80){\vector(1,-1){25}}
	\put(90,90){\circle{20}}\put(84,87){$X_3$}
	\put(85,80){\vector(-1,-1){25}}
	\put(50,50){\circle{20}}\put(46,47){$S$}
	\put(50,40){\vector(0,-1){20}}
	\put(50,10){\circle{20}}\put(46,5){Y}
	\put(10,50){\circle{20}}\put(4,47){$X_1$}
	\put(15,40){\vector(1,-1){25}}
\end{picture}
\caption{Predictive model in which a summary score is used to capture the effect of multiple variables.}
\label{fig:predscore}
\end{figure}

Predictive models often use credit scores, crime scores and other scores as substitutes for economic, demographic and environmental factors. The idea is that a single explanatory summary score can replace multiple explanatory variables for which data is unavailable. Essentially, the collection of omitted variables are related to the summary score and become irrelevant in predicting an outcome if the summary score is included as a predictor. It is said that an outcome is independent of the omitted variables conditional on a valid summary score. This is written as $Y\cip(X_2,X_3)\cd S$, where an outcome $Y$ is explained by the variables $(X_1,X_2,X_3)$ and a summary score $S$ is a valid substitute for the subset of explanatory variables $(X_2,X_3)$. This can be represented by the graph in Figure~\ref{fig:predscore}.

Figure~\ref{fig:predscore} intuitively expresses the assumptions required for $S$ to be a valid summary score. If it is violated then $S$ is not a valid substitute for $X_2$ and $X_3$. The graphical model provides an intuitive representation of the requirement for a summary score. The graph could be used to argue intuitively that there is no direct relation between the omitted variables and the outcome, without quantitative details on the form of relationships. This is quite useful since it is often the only way of justifying the validity of a summary score. Alternative forms of justification are not usually possible because a lack of data on the omitted variables is usually the reason for interest in the summary score in the first place. If data on $S$ were available then formal statistical tests of the conditional independence $Y\cip(X_2,X_3)\cd S$ could be done. For example, if the coefficients of $(X_2,X_3)$ were not statistically significant in a linear model relating $Y$ to $(X_1,X_2,X_3,S)$ then this would suggest the conditional independence was satisfied.

\subsection{Stochastic claims reserving}\label{sec:stochBF}
\begin{figure}[htbp]
	\centering
\begin{picture}(100,100)
	\put(10,90){\circle{20}}\put(4,87){$\alpha$}
	\put(15,80){\vector(1,-1){25}}
	\put(90,90){\circle{20}}\put(84,87){$\beta$}
	\put(85,80){\vector(-1,-1){25}}
	\put(50,50){\circle{20}}\put(44,47){$\phi$}
	\put(50,40){\vector(0,-1){20}}
	\put(50,10){\circle{20}}\put(46,5){C}
	\put(90,50){\circle{20}}\put(84,47){$\tau$}
  \put(85,40){\vector(-1,-1){25}}
	\put(10,50){\circle{20}}\put(4,47){$\psi$}
	\put(15,40){\vector(1,-1){25}}
\end{picture}
\caption{Stochastic Bornhuetter-Ferguson model for ultimate claims $S$, where $\phi$ and $\tau$ are vectors of triangle rows and columns paramaters, $\psi$ is an over-dispersion parameter and $(\alpha,\beta)$ are hyper-parameters.}
\label{fig:stochBF}
\end{figure}
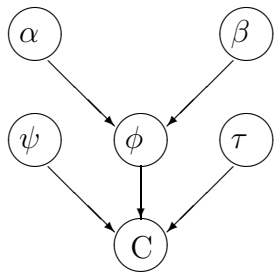

A stochastic version of the Bornhuetter-Ferguson method of claims reserving is described in \citet{V:2008}. The approach describes a hierarchical model in which the distribution of ultimate losses $C$ is defined by vectors of parameters for the rows $\phi$ and columns $\tau$ of an incremental loss triangle. The model assumes losses follow a distribution with an over-dispersion parameter $\psi$. The apriori opinion on the losses for each row of the triangle is expressed as a prior distribution for the row parameters $\phi$ with the parameters $(\alpha,\beta)$, called hyper-parameters. The model is represented by Figure~\ref{fig:stochBF} and implicitly contains the assumption $C\cip (\alpha,\beta)\cd (\phi,\tau,\psi)$. The variables are not deterministic functions of the parameters but their distributions are influenced by the parameters. These are stochastic relationships which are represented by the arrows.

The model in Figure~\ref{fig:stochBF} can be viewed as being built in two stages. The first involves expressing an apriori opinion about $\phi$ via $(\alpha,\beta)$ whereas the second involves predicting the losses $C$ from $(\phi,\tau,\psi)$. Since the row parameters $\phi$ are not influenced by the loss data $C$ then the first stage of conjuring an apriori opinion can be done using external loss data, as is common practice.

The column parameters $\tau$ do not have any arrows leading to them from other parameters since they have a degenerate distribution with constants as parameters. This section assumes parameters are variables with a probability distribution, which is a Bayesian specific philosophy and requires a Bayesian approach to adopt the model in Figure~\ref{fig:stochBF}. The model also implicitly assumes some of the parameters and hyper-parameters are independent $(\phi,\alpha,\beta)\cip\tau\cip\psi$.

Another example of the use of directed graphs in loss reserving is given in \citet{P:2012}. It uses a Bayesian network to express a Kalman filter model of \citet{DZ:1983}.

\subsection{Dependency in capital modelling}
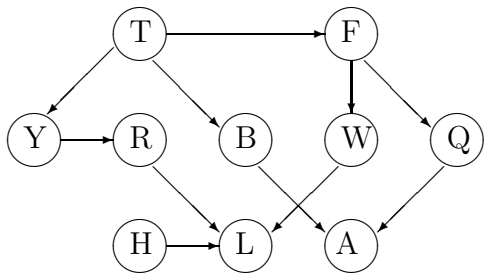
\begin{figure}[htbp]
	\centering
	\begin{picture}(140,100)
	\put(90,10){\circle{20}}\put(86,7){L}
	\put(130,10){\circle{20}}\put(124,7){A}
  \put(125,40){\vector(-1,-1){25}}\put(165,40){\vector(-1,-1){25}}
	\put(95,40){\vector(1,-1){25}}\put(55,40){\vector(1,-1){25}}
	\put(60,10){\vector(1,0){20}}\put(20,50){\vector(1,0){20}}
	
	\put(10,50){\circle{20}}\put(6,47){Y}
	\put(50,50){\circle{20}}\put(46,47){R}
	\put(50,10){\circle{20}}\put(46,7){H}
	\put(90,50){\circle{20}}\put(86,47){B}
	\put(130,50){\circle{20}}\put(126,47){W}
	\put(170,50){\circle{20}}\put(166,47){Q}
		
	\put(50,90){\circle{20}}\put(46,87){T}
	\put(55,80){\vector(1,-1){25}}\put(40,85){\vector(-1,-1){25}}\put(60,90){\vector(1,0){60}}
	\put(130,90){\circle{20}}\put(126,87){F}
	\put(135,80){\vector(1,-1){25}}\put(130,80){\vector(0,-1){20}}
\end{picture}
\caption{Graphical model translation and extension of a capital model example in \citet{SSS:2012}, with economic, underwriting, credit and catastrophe risk factors.}
\label{fig:capmodex}
\end{figure}

Risk aggregation and scenario generators in capital models require some specification of the inter-dependency of risks. Directed graphs allow these to be expressed intuitively. Figure~\ref{fig:capmodex} extends an example from \citet{SSS:2012} and translates it into a graphical model. The variables in the model are interest rate ($T$), cost inflation ($F$), bond price index ($B$), attritional underwriting losses ($W$), measure of softness in the marker to indicate the point in the underwriting cycle ($Y$), equity price index ($Q$), hurricane/catastrophe losses ($H$), reinsurer default ($R$), total liabilities ($L$), total assets ($A$).

Unlike the causal loop diagrams proposed in \citet{SSS:2012}, the relationships between risks in Figure~\ref{fig:capmodex} are stochastic and thus more realistic. The diagram can be intuitively discussed with management before overlaying any technical statistical details. Qualitative judgment about dependency can be done before introducing the complications of  choosing quantitative measures such as copulas and correlation matrices. For example, the omission of a direct link from $T$ to $A$ in Figure~\ref{fig:capmodex} would be based on the opinion that interest rates only influence asset values by influencing bond prices and cost inflation.

The qualitative step could reduce complexities in the quantitative stage because measures of dependency only need to be specified between individual variables and their direct influence. For example, the distribution of $Y$ only needs to be specified conditional on interest rates. Also, only the marginal distribution of a catastrophe loss is specified without any consideration about dependency on the other variables. As with these types of models, the individual relations are aggregated to global connections.

The model encodes probabilistic relations so it is straightforward to express uncertainty about reinsurance default, catastrophe losses or other potential events. The impact on loss distributions, asset values or other consequences are also readily extracted. For example, the model can be queried for the effect of a catastrophe on capital by using inference algorithms to update the distribution of assets and liabilities with the hypothetical evidence that a catastrophe occurred. Predicted capital would then be some quantile or other function of the updated distribution.

While \citet{SSS:2012} refers to models of dependency as `causal', this discussion intentionally avoids that terminology. Causal relations can be expressed with the graphical models but require additional complex semantics beyond the scope of the current discussion. Issues relating to spurious relations can still be represented within the non-causal models with a common variable influencing multiple variables.

\section{Dynamic querying of a network of sensors}\label{sec:dynpred}
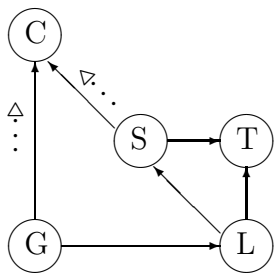
\begin{figure}[htbp]
\centering
\begin{picture}(100,105)
	\put(10,95){\circle{20}}\put(6,92){C}	
	\put(10,15){\circle{20}}\put(6,12){G}
	\put(50,55){\circle{20}}\put(46,52){S}
	\put(90,55){\circle{20}}\put(86,52){T}
	\put(90,15){\circle{20}}\put(86,12){L}
	\put(10,25){\vector(0,1){60}}
	\put(20,15){\vector(1,0){60}}
	\put(40,60){\vector(-1,1){25}}
	\put(60,55){\vector(1,0){20}}
	\put(90,25){\vector(0,1){20}}
	\put(80,20){\vector(-1,1){25}}
	\put(25,70){\rotatebox{135}{\rarr}}\put(0,50){\rotatebox{90}{\rarr}}
	\end{picture}
\caption{Claim alert system linked to various sensors in a home.}
\label{fig:smokcl}
\end{figure}

Graphical models intuitively represent complex structures of relationships. Unlike classic predictive models with a single response and predictors, it incorporates the inter-relationships of predictors and their own predictors. This is helpful in modelling a network of interconnected sensors and the forecasts extracted from their input. Once the model is specified, the network can be updated and queried with fluctuating sensor input as changing evidence. Forecasts are dynamically adjusted from the recurrent network queries. Consider the example in Figure~\ref{fig:smokcl}, which aims to predict whether a fire damage claim arises in a home, using input from various sensors across time. It models the relationship between occurrence of a claim ($C$), temperature of a house ($T$), presence ($S=1$) or absence ($S=0$) of smoke in the house, lights being on in the kitchen ($L$) and the presence ($G=1$) or absence ($G=0$) of the car in the garage.

The model in Figure~\ref{fig:smokcl} is formulated from intuitive reasoning. A claim is more likely to occur if smoke is present and the car is not in the garage, since someone is less likely to be at home. The lights are more likely to be on in the kitchen if the car is in the garage. An extremely high temperature from a fire is more likely if there is smoke and the lights are not on in the kitchen. The chance of a high temperature depends on whether there is smoke and the lights are on in the kitchen, regardless of whether the car is in the garage. The important factor is whether someone is in the kitchen, regardless of whether someone is at home. This collection of assumptions is used to specify the presence and omission of the arrows between variables in Figure~\ref{fig:smokcl}.

At each point in time, the chance of a claim can be updated by the evidence from the observed values of the sensors. For example, if smoke is detected and the car is in the garage then the distribution of $C$ can be updated with the evidence that $S=1$ and $G=1$. An alert could be issued if the mean or some other function of the updated distribution of $S$ crosses a threshold. The dotted arrows in Figure~\ref{fig:smokcl} show the flow of information from evidence. The model also allows alternative queries since each variable may act as a predictor or response. For example, if the smoke alarm input is unavailable then the presence of smoke can be predicted, whereas it was previously used as a predictor. If the temperature sensor input is available then temperature can be used to update the network. The presence of smoke can be inferred if the updated chance of $S=1$, obtained by querying the network, crosses some threshold. As the temperature varies with time, inference about smoke varies accordingly.

The traditional message passing algorithms described in Section~\ref{sec:grmod} can be used to make inference by updating the network with evidence, particularly if the data from the sensor network is stored and used offline. However these have been extended to develop robust algorithms for use in real time processing for networks with asynchronous and noisy readings from sensors that have varying computational capability \citep{CP:2003,CCFIMWW:2006}. Clearly, the dynamic queries are useful for real time detection of triggers. However, actuarial models are more aimed at forecasting whether the trigger leads to a claim. The former and latter are different problems. Even though claim detection is of more importance, data on claims may be sparse and provide predictions with lower accuracy.

\section{Recurrent probabilistic learning}\label{sec:firemod}
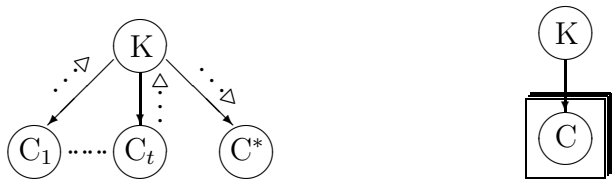
\begin{figure}[htbp]
	\centering
\begin{picture}(100,60)
	\put(10,10){\circle{20}}\put(4,7){C$_1$}
	\put(50,10){\circle{20}}\put(44,7){C$_t$}
	\put(22,7){$\cdots$}\put(24,7){$\cdots$}
  \put(90,10){\circle{20}}\put(84,7){C$^*$}
  \put(40,45){\vector(-1,-1){25}}\put(50,40){\vector(0,-1){20}}\put(60,45){\vector(1,-1){25}}
  \put(50,50){\circle{20}}\put(46,46){K}
	\put(15,30){\rotatebox{45}{\rarr}}\put(55,20){\rotatebox{90}{\rarr}}\put(70,40){\rotatebox{-45}{\rarr}}
\end{picture}\hspace{2cm}
\begin{picture}(100,60)
	\put(50,15){\circle{20}}\put(46,12){C}
	\put(50,45){\vector(0,-1){20}}
  \put(50,55){\circle{20}}\put(46,51){K}
	\squar[30]{35}{65}{0}{30}
	\multiput(65,30)(1,1){3}{\line(0,-1){30}\line(-1,0){30}}
\end{picture}
\caption{Model for predicting a future claim $C^*$ from past claims $C_1,\ldots,C_t$ via a time constant underlying risk characteristic $K$, with classic (left) and plate (right) representations.}
\label{fig:bayes1par}
\end{figure}

All of the models in Section~\ref{sec:dynpred} consisted of relationships among variables at the same time point. Each query only involved computations from evidence within a single time slice. This section discusses models with relationships or dependencies across time. Such models allow queries on variables at a future time point based on evidence from historical time points. They also include static variables that do not vary across time. Consider an example of a model in which $C_t$ is an indicator of whether there is a fire damage claim at time $t$ and $K$ is some measure of the combustibility of a building. The variable $C_t$ is dynamic since claims occur across time. Time may be discrete or continuous like daily intervals or real time respectively. It is assumed that $K$ is constant across time and the claims are influenced by this underlying static risk characteristic. This idea is intuitively captured by the model in Figure~\ref{fig:bayes1par} (left).

Figure~\ref{fig:bayes1par} (left) is formulated intuitively by assuming that claims occur randomly across time for a given combustibility but historical claims are indicative of future claims. These assumptions are written as $C_{i}\cip C_{j}\cd K$ and $C_{i}\ncip C_{j}$ for all $i\neq j$. If data on $K$ is available then future claims can be readily predicted with $\p(C^*\cd K)$ since knowledge of $K$ makes claim history irrelevant.

If data on $K$ is unavailable then claim prediction relies on using claim history to infer or learn about $K$.
At time $t$, the claim history, $C_1,\ldots,C_t$, is used as evidence to update the uncertainty about the level of combustibility of a building from $\p(K)$ to $\p_t(K)$, where $\p_t(\cdot)=\p(\cdot\cd C_1,\ldots,C_t)$. The updated distribution of $K$ can then be used to improve predictions on future claims beyond time $t$. This is a process of learning since forecasts of future claims are dynamically adjusted as claims data is collected. The learning process is represented by the dotted arrows in Figure~\ref{fig:bayes1par} (left). At time $t$, the steps in the algorithm for exact computation are
\begin{enumerate}
\item update: $\p_t(K)\propto\p_{t-1}(K)\times\p(C_t\cd K)$,
\item predict: $\p_t(C^*)=\sum_K\p(C^*\cd K)\p_t(K),$
\end{enumerate}
since $C_{t}\cip(C_1,\ldots C_{t-1})\cd K$. The first step is sequential learning about construction by a recursive update of uncertainty about $K$. It uses the latest claim $C_t$ to update the belief about $K$ at the previous time point, $\p_{t-1}(K)$. The updates are done efficiently with $\p(C_t\cd K)$ instead of $\p_{t-1}(C_t\cd K)$ and thus do not use the full history of claims. This efficiency is only possible because the network structure shows that all dependence between claims is via the construction class. The second step forecasts future claims, $C^*$, using the latest update of knowledge, $\p_t(K)$. Here again efficiency is achieved by exploiting the network structure.

If it were the case that past claims influence future claims then $C_{t}\ncip \{C_{1},\ldots,C_{t-1}\}\cd K$ and the network would have direct arrows between claims in Figure~\ref{fig:bayes1par} (left). The learning algorithm would have to be adjusted by replacing $\p(C_n\cd K)$ with $\p_{n-1}(C_n\cd K)$ and $\p(C^*\cd K)$ with $\p_n(C^*\cd K)$ since claim history is relevant for predictions. The computation still exploits the network structure as much as possible but it does not permit any further efficiency gains. This type of model is similar to an auto-regressive time series model where past realizations of a process influence its future path.

The diagram on the right in Figure~\ref{fig:bayes1par} is a compact representation of the model on the left. It uses the `plate' representation of WinBUGS \citep{LTBS:2000} for repeated instances of the same variable across time. The variable $C_t$ varies with time so is drawn within the plate but $K$ does not vary across the time points so it is drawn outside of the plates. All variables in Figure~\ref{fig:smokcl} vary with time so it could have been drawn entirely within a plate to make the time variation explcit.

The variable $K$ could be a real characteristic such as construction class, which is unavailable for the particular risk being analyzed. Similar applications occur in the medical literature where an unknown underlying disease is diagnosed based on the observable symptoms \citep{FSMB:1989}. Alternatively, the variable $K$ could be an unmeasurable artificial variable such as a score, classifying combustibility according to claim history. It could have even been used for the sole purpose of explaining the dependencies between claims, without representing any real world quantity. Similar models are used in natural language processing to group words into `topics' based on their occurrence in a collection of documents \citep{BNJ:2003}. The topics are inferred groupings or clusters, not pre-defined real variables. From a Bayesian perspective, $K$ could also be a parameter of the claim distribution that is updated with claim history, similarly to the stochastic reserving approach of Section~\ref{sec:stochBF}. In the life insurance literature, related ideas have involved the use of dynamic copulas to capture time varying mortality dependence \citep{CKL:2017}. A conditional distribution was specified for the observations given a time varying parameter that followed a linear additive time series model.

\section{Claims in smart homes with complex interconnections}\label{sec:dynhomeex}
This section proposes examples of the types of models that are useful for modelling claims in a connected home. Their suitability arises because a connected home typically has a network of sensors and other input variables with a complex structure of interconnecting relationships. It is important for a loss model to represent the structure of interconnections to improve predictions and optimize the efficiency of intensive computations. Section~\ref{sec:realclaimalert} focuses on dynamic querying with static and dynamic input, whereas Section~\ref{sec:lmlatentexpl} focuses on learning with hidden risk characteristics.

\subsection{Loss model with dynamic and static risk factors}\label{sec:realclaimalert}
\begin{figure}[htbp]
\centering
	\begin{picture}(140,180)
	\put(90,95){\circle{20}}\put(86,92){C}	
	\put(60,60){\vector(1,1){25}}\put(46,52){P}	
	\put(60,95){\vector(1,0){20}}\put(46,92){K}	
	\put(60,130){\vector(1,-1){25}}\put(46,132){F}	
	\put(90,165){\vector(0,-1){60}}\put(100,170){\vector(1,-1){25}}\put(84,170){M}	
	\put(20,120){\vector(2,1){20}}\put(20,110){\vector(2,-1){20}}\put(6,112){Y}	
	\put(20,80){\vector(2,1){20}}\put(20,70){\vector(2,-1){20}}\put(7,72){I}		
		
	\put(90,15){\circle{20}}\put(86,12){G}
	\put(130,55){\circle{20}}\put(126,52){S}
	\put(170,55){\circle{20}}\put(166,52){T}
	\put(170,15){\circle{20}}\put(166,12){L}
	\put(90,25){\vector(0,1){60}}
	\put(100,15){\vector(1,0){60}}
	\put(120,60){\vector(-1,1){25}}
	\put(140,55){\vector(1,0){20}}
	\put(170,25){\vector(0,1){20}}
	\put(160,20){\vector(-1,1){25}}
	
	\multiput(50,55)(0,40){3}{\circle{20}}\multiput(10,75)(0,40){2}{\circle{20}}\put(90,175){\circle{20}}
	\multiput(130,135)(0,40){1}{\circle{20}}\multiput(126,132)(0,45){1}{B}
	\multiput(130,95)(40,0){1}{\circle{20}}\multiput(124,90)(40,5){1}{W}
	\put(120,95){\vector(-1,0){20}}
	
	\squar[150]{70}{220}{0}{150}
	\multiput(220,150)(2,2){3}{\line(0,-1){150}\line(-1,0){150}}
\end{picture}
\caption{Claim model including dynamic and static variables.}
\label{fig:claimalert}
\end{figure}
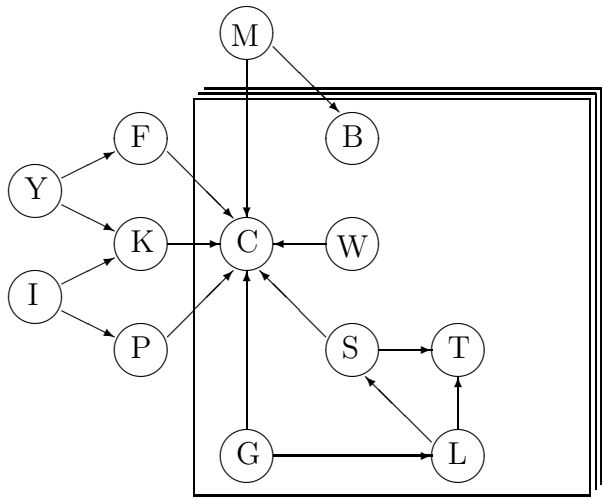

Loss models may supplement input from dynamic sensors with static variables that measure risk characteristics which do not vary with time. Consider the example in Figure~\ref{fig:claimalert}. The dynamic component combines the network in Section~\ref{sec:dynpred}, to detect fire claims, with input from a burglar alarm ($B$) to detect crime and weather station ($W$) to measure rainfall/humidity for weather related losses. These dynamic inputs are linked to the static risk characteristics of the home, which include income class ($I$), year built ($Y$), construction class ($K$), protection class ($P$), flood score ($F$) and crime score ($M$).

The network of interconnections may be formulated based on judgment or learnt from external data. For example, it may be intuitively assumed that at any given point in time, the occurrence of a claim is only related to the triggering of the burglar alarm via their common influence by the crime score of the area. This is represented in Figure~\ref{fig:claimalert} and can be expressed as $C_t\cip B_t\cd M$, where the subscript $t$ indicates that a variable is dynamic and varies with time. Statistical techniques could also have been used to justify the conditional independence $C_t\cip B_t\cd M$ from data. Additionally, the conditional distributions would need to be specified using either approach. The relationships with income and construction and other static variables can be inferred in a similar manner.

Claims are forecast using the model in Figure~\ref{fig:claimalert} by updating the distribution of $C$ with evidence obtained from the input values of the other factors. For example, the fluctuating input from temperature or weather sensors can be propagated throughout the network to update the chance of a claim occurring. Static factors are usually considered as constant over a period, such as a policy year, so evidence from their observed value is not propagated very often. It is even possible to incorporate dynamic sensor data as a static variable. For example, an alternative model could move $B_t$ from the plate and recode it as some time constant variable $B$, which summarizes the burglar alarm trigger frequency across a year.

\subsection{Loss model with structured latent explanatory factors}\label{sec:lmlatentexpl}
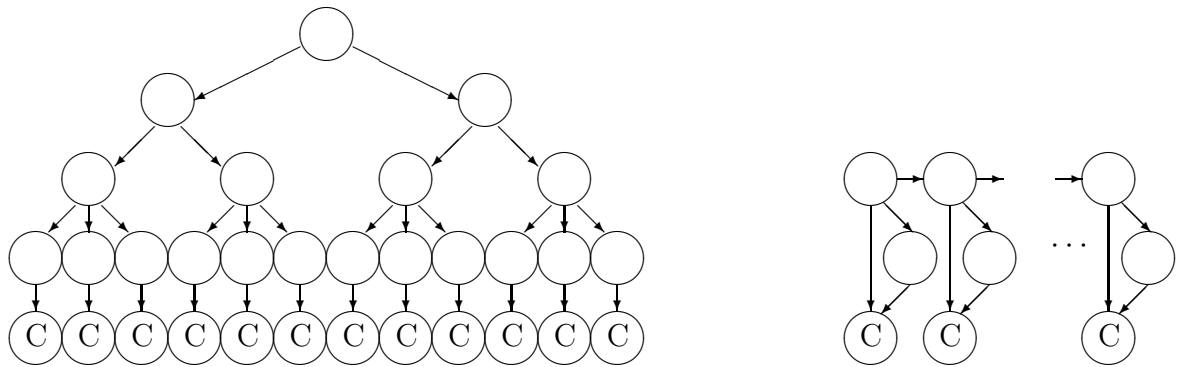
\begin{figure}[htbp]
\centering
\begin{picture}(255,130)
	\multiput(15,10)(20,0){12}{\circle{20}}
	\multiput(15,40)(20,0){12}{\circle{20}}
	\multiput(35,70)(60,0){4}{\circle{20}}
	\multiput(65,100)(120,0){2}{\circle{20}}
	\put(125,125){\circle{20}}
			
	\multiput(30,60)(60,0){4}{\vector(-1,-1){10}}
	\multiput(40,60)(60,0){4}{\vector(1,-1){10}}
	\multiput(35,60)(60,0){4}{\vector(0,-1){10}}
	\multiput(15,30)(20,0){12}{\vector(0,-1){10}}
	
	\multiput(60,90)(120,0){2}{\vector(-1,-1){15}}
	\multiput(70,90)(120,0){2}{\vector(1,-1){15}}
	
	\put(115,120){\vector(-2,-1){40}}
	\put(135,120){\vector(2,-1){40}}
		
	\multiput(11,7)(20,0){12}{C}
\end{picture}\hspace{2cm}
\begin{picture}(150,90)
	\multiput(15,10)(30,0){2}{\circle{20}}
	\multiput(30,40)(30,0){2}{\circle{20}}
	\multiput(15,70)(30,0){2}{\circle{20}}
	
	\multiput(25,70)(30,0){3}{\vector(1,0){10}}
	\multiput(15,60)(30,0){2}{\vector(0,-1){40}}
	\multiput(20,61)(30,0){2}{\vector(1,-1){11}}
	\multiput(30,30)(30,0){2}{\vector(-1,-1){11}}
	
	\multiput(11,7)(30,0){2}{C}
	
	\put(105,10){\circle{20}}
	\put(120,40){\circle{20}}
	\put(105,70){\circle{20}}
	\put(105,60){\vector(0,-1){40}}
	\put(110,61){\vector(1,-1){11}}
	\put(120,30){\vector(-1,-1){11}}
	\put(101,7){C}\put(83,42){$\cdots$}
\end{picture}
\caption{Examples of tree model for weather related claims (left) and emission model for fire related claims (right).}
\label{fig:claimalerth}
\end{figure}

Loss models which represent complex interconnections are also useful for forecasting claims when data on some of those variables remain unavailable. Although the structure of connected variables without data is hidden, evidence on the observed variables can still be propagated via its hidden nodes and links to facilitate predictions. The structure may be formulated from scientific hypotheses about the hidden mechanisms or inferred from external data, possibly using missing data techniques such as the EM-algorithm. Applications of such models in bioinformatics, language and speech processing, image processing and spatial statistics are described in \citet{WJ:2008}. If data on $K$ was unavailable for the model in Section~\ref{sec:firemod} then that would be a simple example of these latent variable models. Further examples are given in Figure~\ref{fig:claimalerth}, where $C$ represents the occurrence of a claim and the hidden explanatory structure consists of the unlabelled variables.

The tree model in Figure~\ref{fig:claimalerth} (left) conjectures that monthly property damage claims, $C$, are weather related and arise from a hierarchy of hidden climate factors. An annual factor influences bi-annual factors, which successively influence quarterly and monthly factors. As data on monthly claim occurrences are accumulated, it can be used as evidence to propagate throughout the network and update beliefs about the underlying climate patterns, for which there is no data. The updated beliefs are in turn useful for forecasting claims. Essentially, evidence from observed historical claims are propagated via the hidden climate factors to update the distribution of claims for future months. The explicit representation of the hidden graph structure is important since it is exploited by the propagation algorithms for efficiency. A special case of these algorithms is used in inferring phylogenetic trees by computing the likelihood of various evolutionary paths for protein sequences \citep{F:1981}. The latent factors in Figure~\ref{fig:claimalerth} (left) could have represented criminal activity or other risk factors that follow the same hierarchy of seasonality. However, such a model would require knowledge of the mechanisms for the extended risk factors.


The model in Figure~\ref{fig:claimalerth} (right) provides an alternative type of hypothesis for explaining fire related claim occurrences with hidden structures. It is an emission model where the top hidden layer is a recurrent variable representing whether the house/kitchen is occupied and `emits' claims, in the sense that it influences whether a claim occurs. In addition to the transitioning occupancy status, the occurrence of a claim is also influenced by whether any smoke/fire actually occurs at each time point, which is the middle hidden layer. Intuitively, the model assumes that the occurrences of smoke/fire are influenced by the occupancy status but are not directly related to one another across time. Their only relations are via their common influence by the occupancy state of the house. Data on claim occurrences are evidence that can be used to update beliefs about the underlying occupancy pattern and then forecast future claims. The algorithm for inferring occupancy patterns intuitively filters the observed claims data to remove the noise from its relationship with occupancy and remove any distortion of the relationship by smoke/fire occurrences. This type of model is used in speech recognition to infer a sequence or pattern of words from speech signals \citep{HAHR:2001}.


Apart from the example given here, there are many other variants of models with hidden variables in the literature. Some of these can potentially be used in actuarial applications. Although the examples discussed involved real variables that are hidden, it is also possible to define models with hidden structures that are purely mathematical constructs. In such models, the latent structure is included to capture the features of the observable variables. A real world interpretation may be philosophically comforting but is not necessary to apply the technical machinery. An example is in image processing applications, where certain models include recurrent hidden variables that are each only locally connected to small neighbourhoods of pixels \citep{W:2002}. The hidden variables capture dependency among the observed pixel values and are recurrent across the space of pixels, instead of time. Models with complex layers of hidden variables are sometimes referred to as `deep', just as with neural networks consisting of hidden layers.

\section{Discussion}

The graphical models for losses were primarily used for predicting claims from risk factors. However, the same networks and algorithms can be exploited to predict risk factors from claims data. This reversal of evidence flow may assist in the detection of ineffective sensors. Querying a network, that has been updated with evidence of multiple fire related claims, would indicate a high chance of multiple instances of the smoke alarm being triggered. If few triggers were recorded then this unusual deviation of prediction from observation suggests that the smoke alarm is faulty or ineffective at preventing the types of fires that are most common in the building. The same network may include crime related or other types of claims and may also identify ineffective burglar alarms or other sensors. Detecting ineffective loss prevention mechanisms cannot be done on an individual level so it is beneficial that the approach here is available since it uses data on multiple policies.

Using the same approach as detecting faulty sensors, claims data can also be used to identify irregularities in policyholder reported risk characteristics. Again, the process uses the same network and algorithms as developed for forecasting losses. This is helpful for scrutinizing risk factors that may not be verified by the insurer, such as whether a house is occupied frequently. Querying a network, that has been updated with evidence of multiple fire claims and triggered smoke alarms, would indicate a high chance that the house is not often occupied. Otherwise the alerts would likely have notified occupants to minimize damage and prevent claims. If the related policy lists the home as frequently occupied then this unusual inconsistency may warrant further investigation. The loss models are similar to claim alert system that detect the combined result of multiple perils and the approach uses the output to highlight unusual input. The update process does not identify definite fraud since the predictions are subject to uncertainty. Nevertheless, the process of highlighting irregularities can potentially reduce fraud.

For certain examples, the flow of information was shown on the graphs during the process of updating predictions with evidence. These are some of the simpler cases since the updates are not always so straightforward to represent. They may involve transformations of the graphs and computations across subgroups of variables. The update process was also only used for evidence about variables taking some fixed value. However, the inference algorithms are more widely applicable and can deal with uncertain evidence. This type of evidence arises if there is information to indicate that a variable takes a value with a certain probability. For example, a network could be updated with the uncertain evidence from a smoke alarm that there is a 70\% chance of smoke being present. This type of evidence can arise from imprecise sensors as it does not necessarily mean that smoke is present in this case.

A useful feature of the graphs is that they capture qualitative knowledge for easy communication. However the step from knowledge to model is not always as simple as the examples given in this discussion. In general, multiple graphs may represent the same assumptions. For a very simple example, both $A\longrightarrow B \longrightarrow C$ and $A\longleftarrow B \longleftarrow C$ represent the assumption that $A$ and $C$ are related via $B$, written as $C\cip A\cd B$. The directions of the arrows are reversed but represent the same dependency. Therefore knowledge of conditional independence relations does not always translate to a unique graphical model. Further knowledge of the directions of influence could be needed and such issues become worse with more variables and more dependencies. 

The models described throughout are meant to demonstrate the usefulness and capabilities of directed graphs in various actuarial applications. They may require adjustments and extensions for implementation in practice. Any particular analysis could be more appropriately modelled with different interconnections and dependencies or more variables. Other types of graphical models with undirected edges or hybrids can also be useful but directed graphs have gained popularity because of their intuitive appeal and relative simplicity.

\end{document}